\documentclass[10pt,twocolumn,letterpaper]{article}

\usepackage{bm}
\usepackage{iccv}
\usepackage{times}
\usepackage{epsfig}
\usepackage{graphicx}
\usepackage{amsmath}
\usepackage{amssymb}
\usepackage{multirow}
\usepackage{color,soul}
\usepackage[accsupp]{axessibility} % Improves PDF readability for those with disabilities.

\usepackage[pagebackref=true,breaklinks=true,letterpaper=true,colorlinks,bookmarks=false]{hyperref}

\iccvfinalcopy % *** Uncomment this line for the final submission

\def\iccvPaperID{6427} % *** Enter the ICCV Paper ID here

\begin{document}

%%%%%%%%% TITLE
\title{MultiSiam: Self-supervised Multi-instance Siamese Representation Learning for Autonomous Driving}

\author{
    Kai Chen$^{1}$
    \quad
    Lanqing Hong$^{2}$
    \quad
    Hang Xu$^{2}$
    \quad
    Zhenguo Li$^{2}$
    \quad
    Dit-Yan Yeung$^{1}$
    \\
    $^{1}$Hong Kong University of Science and Technology
    \quad
    $^2$Huawei Noah's Ark Lab
    \\
    {\tt\small kai.chen@connect.ust.hk
    \quad
    \{honglanqing, xu.hang, li.zhenguo\}@huawei.com
    \quad
    dyyeung@cse.ust.hk}
}

\maketitle
% Remove page # from the first page of camera-ready.
% \ificcvfinal\thispagestyle{empty}\fi

%-------------------------------------------------------------
%%%%%%%%%%%%%%%%%%%%%%%%%
% Note:
% correct the reference between main body and appendix
%%%%%%%%%%%%%%%%%%%%%%%%%

%-------------------------------------------------------------
%%%%%%%%% ABSTRACT
\begin{abstract}
  Autonomous driving has attracted much attention over the years but turns out to be harder than expected, probably due to the difficulty of labeled data collection for model training.
  Self-supervised learning (SSL), which leverages unlabeled data only for representation learning, might be a promising way to improve model performance.
  Existing SSL methods, however, usually rely on the single-centric-object guarantee, which may not be applicable for multi-instance datasets such as street scenes.
  To alleviate this limitation, we raise two issues to solve: (1) how to define positive samples for cross-view consistency and (2) how to measure similarity in multi-instance circumstances.
  We first adopt an IoU threshold during random cropping to transfer global-inconsistency to local-consistency.
  Then, we propose two feature alignment methods to enable 2D feature maps for multi-instance similarity measurement.
  Additionally, we adopt intra-image clustering with self-attention 
  for further mining intra-image similarity and translation-invariance.
  Experiments show that, when pre-trained on Waymo dataset, our method called \textbf{\textit{Multi}}-instance \textbf{\textit{Siam}}ese Network (MultiSiam) 
  remarkably improves generalization ability and 
  achieves state-of-the-art transfer performance on autonomous driving benchmarks, including Cityscapes and BDD100K, while existing SSL counterparts like MoCo, MoCo-v2, and BYOL show significant performance drop.
  By pre-training on SODA10M, a large-scale autonomous driving dataset, MultiSiam exceeds the ImageNet pre-trained MoCo-v2, demonstrating the potential of domain-specific pre-training.
  Code will be available at \url{https://github.com/KaiChen1998/MultiSiam}.

\end{abstract}

%-------------------------------------------------------------
%%%%%%%%% BODY TEXT
%%%%%%%%% INTRODUCTION
\vspace{-1mm}
\section{Introduction}\label{sec:intro}

\begin{figure}[t]
\begin{center}
  \includegraphics[width=1.0\linewidth]{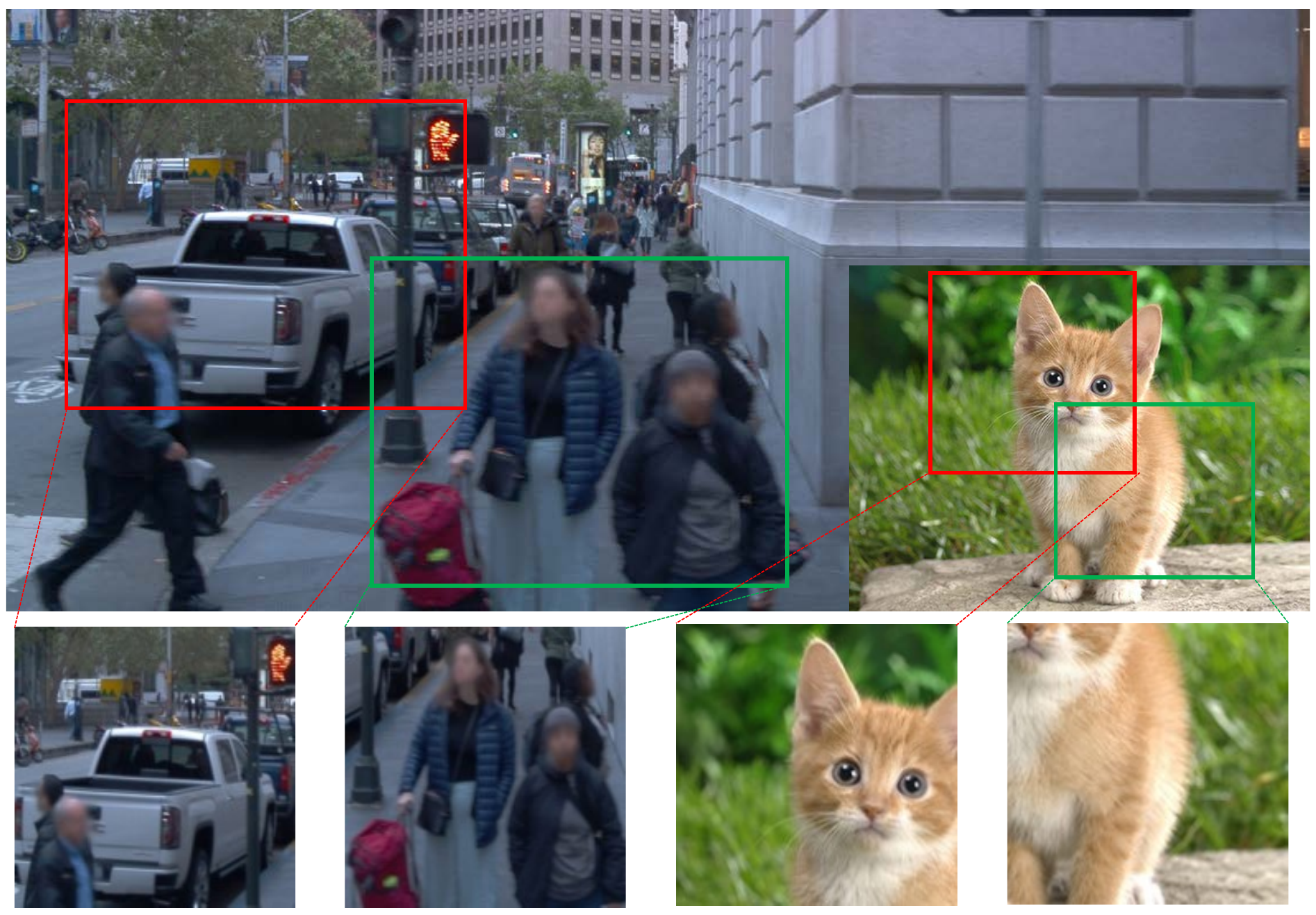}
\end{center}
\caption{{\bf Visualizations of different random crops on Waymo~\cite{sun2020scalability} (left) and ImageNet \cite{deng2009imagenet} (right).}
On ImageNet, images are small and pre-processed to guarantee only one object in the center part of it (\ie, single-centric-object) mostly.
However, images from Waymo are of high resolution and contain multiple instances.
Different random crops might represent different semantic meanings (\ie, global-inconsistency), which will restrict the effectiveness of current self-supervised learning methods.}
\label{fig:inconsistency}
\vspace{-3mm}
\end{figure}

%-------------------------------------------------------------

Autonomous driving has attracted much attention over the years \cite{pendleton2017perception,janai2020computer,yurtsever2020survey}. 
However, it becomes harder than people have expected in such an age of AI advances.
Fully autonomous cars are still out of reach except in special trial programs, mainly due to the limitation of model performance. 
One of the main restrictions is that the annotation cost of self-driving datasets is much more expensive than other datasets.
Considering that autonomous cars keep collecting unlabeled data when operating, self-supervised learning (SSL) might be a promising way to ease the desire for labeled data and improve model performance, which has achieved remarkable transfer results on different downstream tasks using unlabeled data only.

Existing SSL methods are mainly based on the pretext called instance discrimination and cross-view consistency framework, whose basic assumption is that different views (\eg data augmentation) of a single image should be consistent in the feature space under different metrics, such as 
cosine distance \cite{grill2020bootstrap,chen2020exploring}, 
clustering assignments \cite{caron2020unsupervised}
and discriminability from negative samples \cite{wu2018unsupervised,tian2019contrastive,chen2020simple,he2020momentum,chen2020improved,misra2020self}, 
This assumption is satisfied well with {\it single-centric-object} datasets such as ImageNet.
In self-driving, nevertheless, the data are usually high-resolution images containing multiple instances on a single image (see Figure~\ref{fig:inconsistency} for an illustration).
Here we define {\it instance} as any individual object regardless of its semantic class following Wu~\etal~\cite{wu2018unsupervised}.
In this case, different crops may correspond to different instances and represent different semantic meanings, which results in the {\it global-inconsistency} of multi-instance images.
The effectiveness of instance discrimination and cross-view consistency can no longer be guaranteed.
% On the other hand, models pre-trained on single-centric-object datasets such as ImageNet may suffer from \textcolor{red}{domain gaps} when fine-tuned on downstream self-driving tasks with multi-instance datasets.

% exiting self-supervised learning methods may suffer from \textcolor{red}{drawback} when pre-training.
% So to some extent, even if current self-supervised learning doesn't require human annotation any more, it still can't scale to more common datasets as we want. It's inappropriate to directly deploy state-of-the-art models on multi-instance datasets.
% But domain gap still exists inevitably for specific datasets (\eg street scenes and autonomous driving).

To adapt current instance discrimination and cross-view consistency framework to multi-instance circumstances, we need to solve two problems: (1) how to define positive samples for cross-view consistency and (2) how to calculate the similarity of two randomly generated views within multi-instance images. 
Considering the locality of images, we first add an IoU threshold during random cropping as a proxy to control the two views not too far from each other and transfer global-inconsistency to local-consistency.
In order to distinguish different instances, we maintain the final 2D feature maps of the backbone networks and propose the {\it RoI alignment} and {\it offset alignment} to solve the feature misalignment introduced by it (see Figure \ref{fig:misalignment}(a)), which is usually neglected when global pooling layers are adopted.
Moreover, we observe a hierarchy of clusters existing in multi-instance circumstances naturally,
so to model the relationships between instances, 
we perform clustering within a single image not for inter-image similarity~\cite{caron2018deep,caron2020unsupervised},
but for further mining intra-image similarity.
To ease the ambiguity of cluster assignments, we deploy a self-attention mechanism with the predictor for more precise cluster prediction.
Translation-invariance is also enhanced in the learned representation, which is beneficial for downstream pixel-level visual tasks like semantic segmentation.

% Instead of taking random crops uniformly in a single image, if the two crops are "close" enough with each other, we can assume locally they are representing similar semantic meanings and should be consistent in the feature space.

% Although there might be multiple instances in a street scene image, the class set is usually small. For example, you might capture multiple pedestrians and cars in a random crop. Since they belong to the same semantic class, they should be more similar compared to others like the background pixels.

The main contributions of this work contain three parts:
\begin{enumerate}
    \item We propose the Multi-instance Siamese Network ({\it MultiSiam}) for extending the cross-view consistency framework to multi-instance circumstances by dealing with positive sample definition and similarity measurement of 2D feature maps. 
    \item Experiments on Cityscapes~\cite{cordts2016cityscapes} and BDD100K~\cite{yu2018bdd100k} show that {\it MultiSiam} pre-trained on Waymo~\cite{sun2020scalability} has
    stronger generalization ability to multi-instance datasets and
    achieves state-of-the-art transfer performance on downstream autonomous driving benchmarks compared with MoCo, MoCo-v2 and BYOL.
    Moreover, {\it MultiSiam} pre-trained on SODA10M~\cite{han2021soda10m}, a large-scale autonomous driving dataset, exceeds the ImageNet pre-trained MoCo-v2, revealing the potential of domain-specific pre-training.
    \item To the best of our knowledge, our work is the first to perform self-supervised learning on large-scale high-resolution multi-instance street scene datasets (\eg Waymo), which will be beneficial for empowering SSL in further autonomous driving research.
\end{enumerate}

%(1) we demonstrate current self-supervised learning methods are still restricted to ImageNet dataset because of the single-centric-object guarantee, and not able to scale to more general multi-instance datasets as originally claimed. 
%No need for annotation doesn't equal to global scalability.

% (2) we solve two problems to extend the cross-view consistency framework to multi-instance datasets and propose a novel model that can make better use of multi-instance datasets
% (3) Extensive experiments on Cityscapes and BDD show that our model pre-trained on Waymo dataset can achieve comparable transfer results on downstream semantic segmentation benchmarks with state-of-the-art models pre-trained on high-quality ImageNet dataset.

% Our work is the first to perform self-supervised learning on large-scale high-resolution multi-instance street scene datasets and will be beneficial for empowering self-supervised learning in further autonomous driving research.

%-------------------------------------------------------------

\begin{figure*}[t]
\begin{center}
  \includegraphics[width=0.8\linewidth]{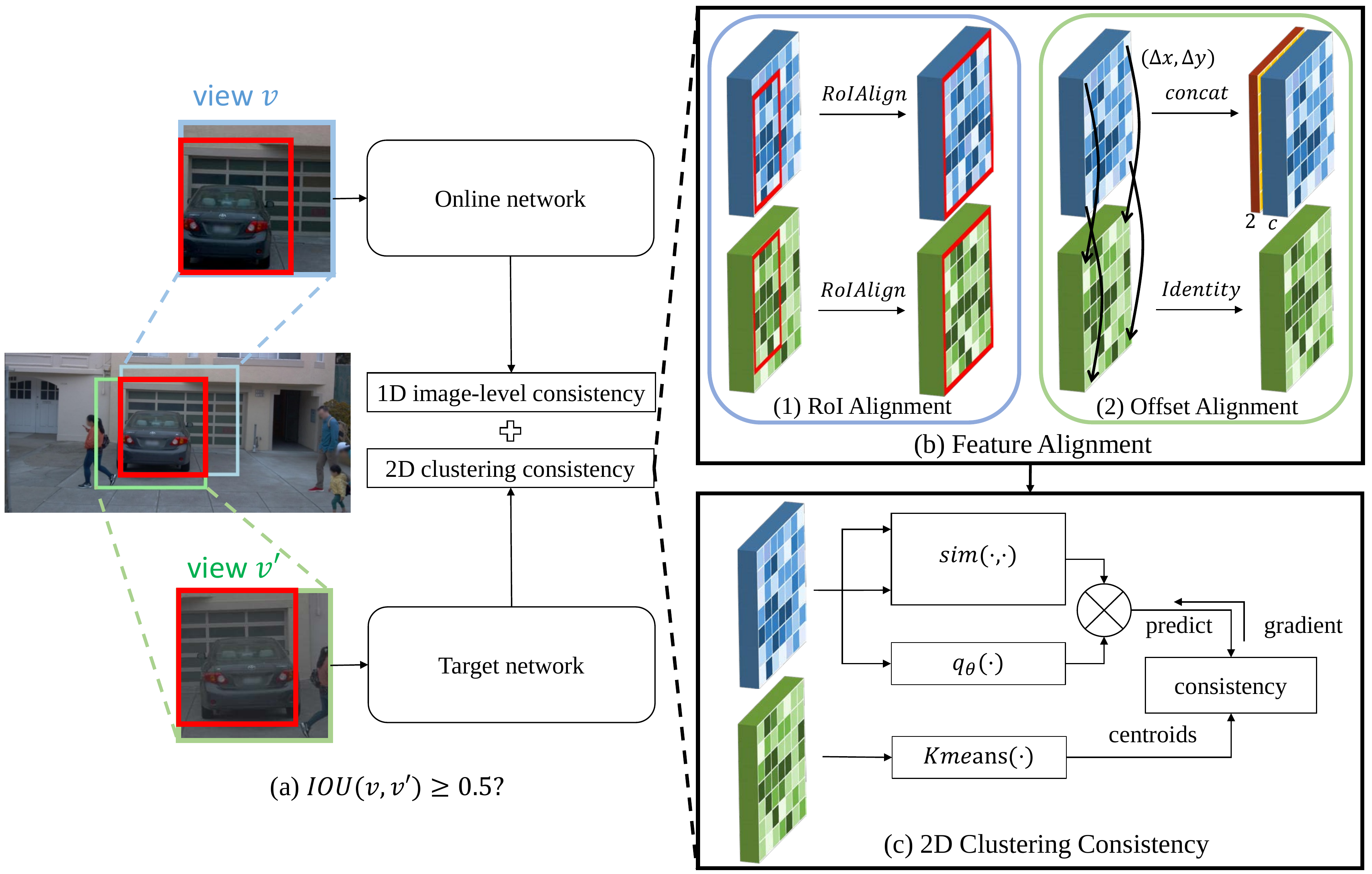}
\end{center}
\caption{\textbf{Model architecture of \textit{MultiSiam}.} 
(a) Two views $v$ and $v^\prime$ are fed for further calculation only when their IoU value is larger than a given threshold.
(b) We maintain the 2D feature maps and propose two methods for feature alignment after projection and flipping back (more details in Figure \ref{fig:misalignment}).
{\it RoI alignment} extracts features of the intersection region only, while {\it offset alignment} provides the coordinate offset map to the predictor for implicit feature alignment.
(c) Intra-image clustering is performed on the aligned target feature map and the online network needs to predict the cluster centroids.
We deploy a self-attention mechanism to deal with the ambiguity of clustering.
}
\label{fig:architecture}
\end{figure*}

%-------------------------------------------------------------

\section{Related Work}\label{sec:related}

\paragraph{Contrastive learning.} 
Contrastive learning is widely used in self-supervised representation learning, which has shown promising performance for various tasks~\cite{wu2018unsupervised,tian2019contrastive,chen2020simple,he2020momentum,chen2020improved,misra2020self,tian2020makes}. 
The main idea is to consider every single image as a separate class and train the model to pull positive sample pairs closer while pushing negative sample pairs away using the InfoNCE loss~\cite{oord2018representation}.
MoCo~\cite{he2020momentum,chen2020improved} proposes to maintain a queue of negative samples with a Siamese network and shift one branch of the network into a momentum encoder to improve the consistency of the queue.
SimCLR~\cite{chen2020simple}, however, directly uses negative samples co-existing in the current batch. 
Contrastive learning requires
comparing each positive with many other negative samples to work well, which is usually achieved by crops from a large amount of single-centric-object images.
In multi-instance data such as street scenes, however, two random crops from one single image may correspond to different instances, leading to the possible ineffectiveness of existing contrastive methods.

\vspace{-3mm}
\paragraph{Clustering.} 
Clustering is another popular paradigm for unsupervised representation learning~\cite{caron2018deep,caron2019unsupervised,asano2019self,caron2020unsupervised}.
The main idea is to consider each cluster as a separate class instead of each single image as contrastive learning does to capture the similarity between images by clustering the representations and learning to predict the cluster assignments alternatively.
Deep Cluster~\cite{caron2018deep} iteratively clusters all the images based on current representations and treats cluster indexes as pseudo labels to train a classifier from scratch. 
Although there is no need for negative samples, a costly offline clustering process is introduced.
SwAV~\cite{caron2020unsupervised} conducts online clustering with a Siamese network by computing the assignment from one branch and predicting it from the other under a balanced partition constraint for each batch.
Different from previous methods, in the proposed {\it MultiSiam}, we perform clustering within a single image instead of the whole dataset to further discover intra-image similarity in multi-instance circumstances.

\vspace{-3mm}
\paragraph{Cosine similarity.} 
Both negative samples and clustering are considered useful to prevent model collapse in self-supervised learning.
BYOL \cite{grill2020bootstrap} instead proposes a cross-view consistency framework with cosine similarity only that can produce meaningful representations with the help of a momentum target network.
SimSiam \cite{chen2020exploring} further points out that the stop-gradient operation is the critical component to prevent model collapse based on an EM-like hypothesis.
Due to its simplicity and effectiveness, we adopt BYOL as our baseline model and show that intra-image clustering is a better similarity metric in multi-instance circumstances.

% \paragraph{(Optional) Traditional unsupervised learning.}
% Traditional unsupervised learning mainly includes generative methods~\cite{kingma2013auto,goodfellow2014generative} and hand-crafted pretext tasks~\cite{doersch2015unsupervised,noroozi2016unsupervised,gidaris2018unsupervised}.

%-------------------------------------------------------------

\section{Method}\label{sec:method}

In this section, we will first introduce the BYOL \cite{grill2020bootstrap} in Section \ref{sec:BYOL},
which can achieve state-of-the-art transfer performance without the need for either negative samples or clustering.
Due to its effectiveness and simplicity, we choose BYOL as our baseline model.
Then we extend the cross-view consistency framework to multi-instance circumstances by discussing two problems: how to define positive samples and how to measure similarity in multi-instance images and propose our final {\it MultiSiam} model in Sections~\ref{sec:positive} and \ref{sec:cluster}, respectively.

%-------------------------------------------------------------

\subsection{Preliminary: BYOL}\label{sec:BYOL}
Given a single image $x$, BYOL learns the representation by maximizing the similarity of two random views $v$ and $v^\prime$ in the feature space.
BYOL first randomly generates two views $v \sim \mathcal{T}(x)$ and $v^\prime \sim \mathcal{T}^\prime(x)$, which are then fed into the online and target networks separately.
The online network consists of a backbone with a global pooling layer, an MLP projector, and an MLP predictor, 
while the target network shares the same architecture except for the final predictor.
To prevent model collapse, BYOL adopts a stop-gradient operation on top of the target network.
Finally, BYOL minimizes the cosine distance between the online network's prediction $q$ and the target network's projected feature $z^\prime$ as the 1D image-level consistency by:
\begin{equation}
    \mathcal{L}_{1D\_img} \triangleq -cos(q, z^\prime) = -\frac{\langle q, z^\prime \rangle}{||q||_2 \cdot ||z^\prime||_2},
\end{equation}
where $cos(\cdot, \cdot)$ is the cosine similarity and $q$ and $z^\prime$ are both 1D feature vectors.

To further improve performance, BYOL updates the parameters of the target network $\xi$ as the exponential moving average of the online network's parameters $\theta$ as:
\begin{equation}
    \xi \leftarrow \tau\xi + (1 - \tau)\theta,
\end{equation}
where $\tau \in [0,1]$ is the momentum and will increase to 1.0 until the end of training.

%-------------------------------------------------------------

\subsection{Positive Samples in Multi-instance Data}\label{sec:positive}
The basic assumption of instance discrimination is that different views of the same image should be consistent in the feature space.
% Based on the cross-view consistency framework, state-of-the-art methods adopt different metrics to calculate feature similarity and prevent model collapse.
However, this strong assumption is only satisfied well on ImageNet \cite{deng2009imagenet} with single-centric-object guarantee but not scalable to more realistic datasets with multiple instances on a single image like Waymo \cite{sun2020scalability}, as shown in Figure \ref{fig:inconsistency}. 
A new definition of positive samples is definitely needed to extend cross-view consistency framework to multi-instance circumstances.

%-------------------------------------------------------------

\begin{figure}[t]
\begin{center}
  \includegraphics[width=0.8\linewidth]{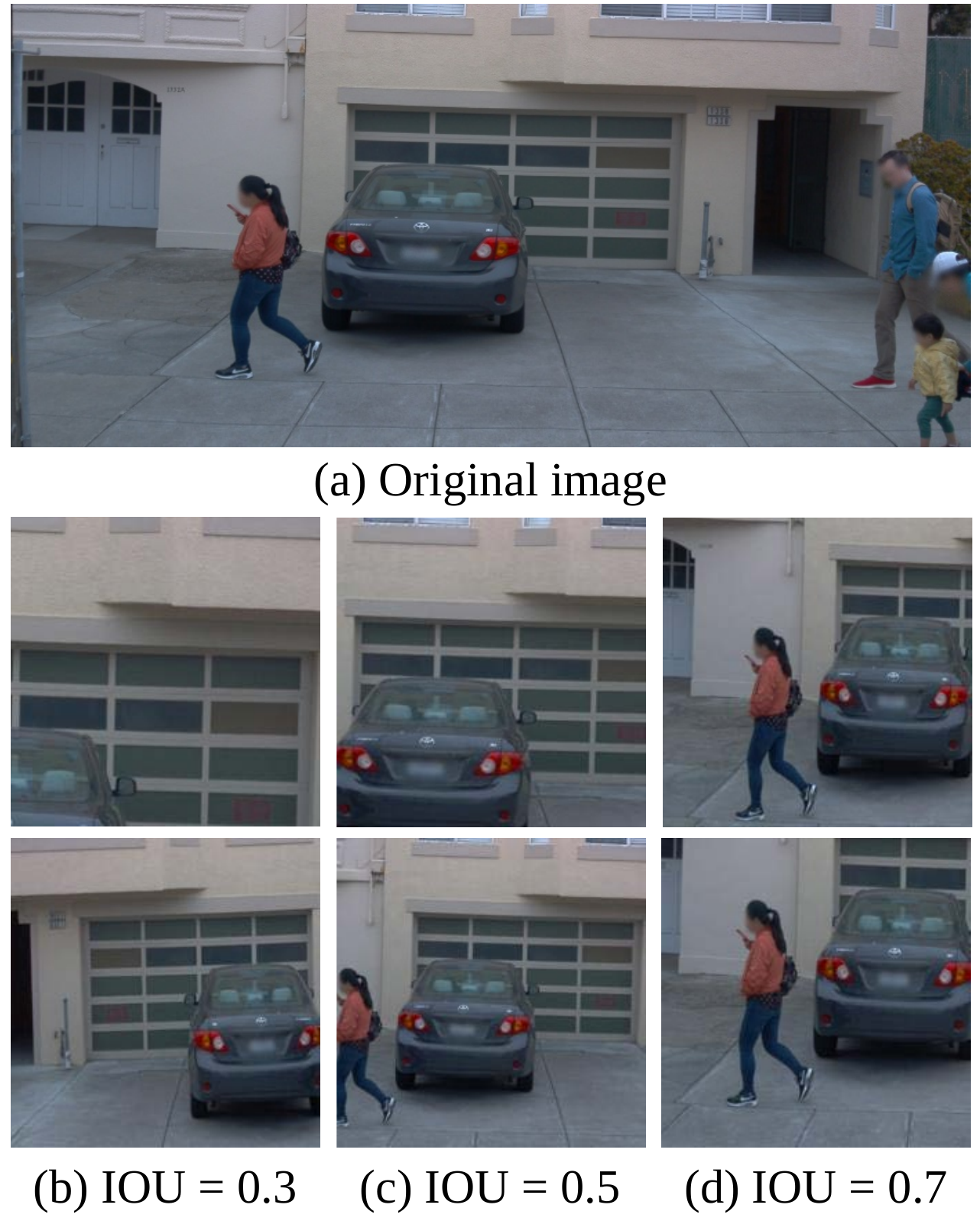}
\end{center}
\vspace{-3mm}
\caption{{\bf Random crops with different Intersection-over-Union values on Waymo (each column).} 
(a) Original image; (b) $IoU=0.3$; (c) $IoU=0.5$; (d) $IoU=0.7$.
When $IoU=0.3$, the two random crops suffer from global-inconsistency.
As the IoU value raises, the two random views are restricted within a local region, and the local-consistency increases.
When $IoU=0.7$, the two views seem nearly the same.
As shown in Section \ref{sec:ablation}, IoU controls a trade-off between noise and data complexity during data augmentation.
}
\label{fig:IoU}
\vspace{-5mm}
\end{figure}

%-------------------------------------------------------------

\paragraph{Intersection-over-Union as a proxy.}
Images are continuous natural signals and have strong locality. 
One of the main reasons why inconsistency happens in multi-instance cases is that the two random crops may be far away from each other.
Considering the locality of images, if the two views are ``close" enough, it is reasonable to assume that they represent the same semantic meanings in a local region and transfer global-inconsistency to local-consistency.

Here we propose to use the Intersection-over-Union $IoU(v, v^\prime)$ as a proxy about how ``close" the two random crops are and set an additional IoU threshold during data augmentation.
Two views are used for further calculation only when their $IoU(v, v^\prime)$ is larger than the pre-defined threshold. 
We show how different IoU values will affect the consistency of the two random crops in Figure \ref{fig:IoU}.
When $IoU(v, v^\prime) = 0.3$, we can still see the car in the bottom crop but the top crop has been mainly occupied by the garage.
As the $IoU(v, v^\prime)$ rises gradually, the semantic meanings of the two crops align better and when $IoU(v, v^\prime) = 0.7$, the two views seem nearly the same.

In Section \ref{sec:ablation}, we will show that using different IoU thresholds is actually a trade-off between noise and data complexity during data augmentation.
If not specified, we set the IoU threshold to be 0.5 in all experiments by default.

%-------------------------------------------------------------

\vspace{-1mm}
\paragraph{Consistency learning via 2D feature map.}

Even with the IoU threshold, there might be multiple instances in a single crop, as shown in Figure \ref{fig:IoU}(d).
State-of-the-art methods all deploy a global pooling layer at the end of the backbones to produce a 1D feature vector, which will lose the spatial and structural information of 2D feature space, and the model cannot distinguish different instances any more.

As a result, we discard the global pooling layers and maintain the final 2D feature maps $F,F^\prime \in \mathcal{R}^{H\times W\times C}$ of both online and target backbone networks.
Meanwhile, we replace the MLPs in the projectors and predictor with $1 \times 1$ convolution layers in order to match the 2D structure while keeping the amount of parameters unchanged.
% The final projector and predictor both consist of a $1 \times 1$ convolution, a synchronized batch normalization, a ReLU activation and a second $1 \times 1$ convolution by default. 

%-------------------------------------------------------------

\begin{figure}[t]
\begin{center}
  \includegraphics[width=1.0\linewidth]{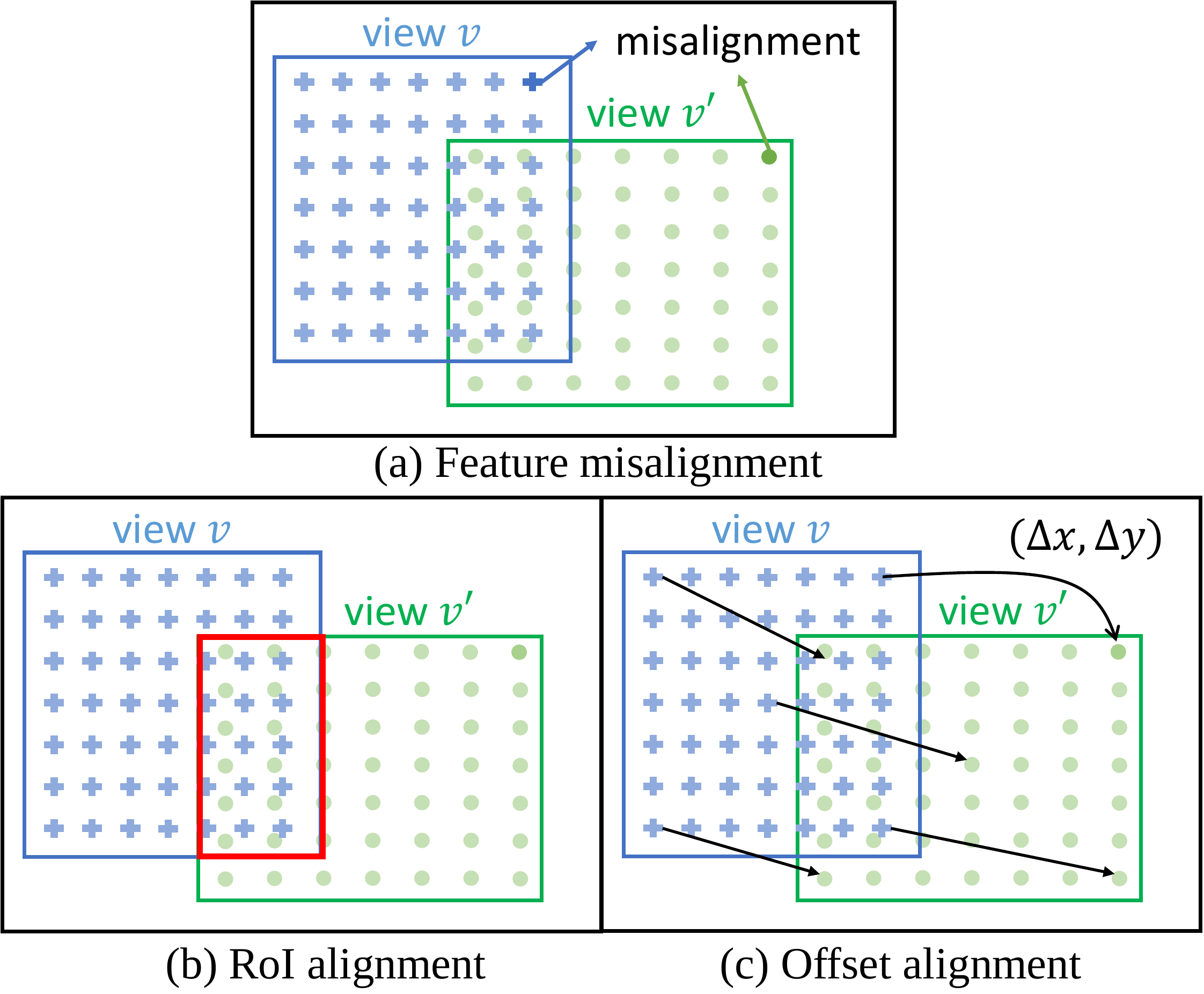}
\end{center}
\vspace{-3mm}
\caption{{\bf Feature misalignment} of two random views (\ie, blue and green boxes) in a local region (\ie, the black box).
(a) Feature misalignment: top right corners of $v$ and $v^\prime$ represent different pixels, so one-to-one correspondence no longer exists; (b) RoI alignment: use RoI Align \cite{he2017mask} to extract features of overlapping regions (\ie, the red box) only; (c) Offset alignment: calculate the coordinate offset of each pixel pair at the same relative position to get the offset map $\Delta C$, which is then concatenated with the projected online 2D feature map $G=g_{\theta}(F)$ and fed into the predictor.}
\label{fig:misalignment}
\vspace{-4mm}
\end{figure}

%-------------------------------------------------------------

\vspace{-3mm}
\paragraph{Feature misalignment.}
However, there is no free lunch.
Using 2D feature maps introduces another feature misalignment problem that has not been taken into account when global pooling layers are available, as shown in Figure \ref{fig:misalignment}(a).
The top right corners of $v$ and $v^\prime$ lie in different positions.
The one-to-one correspondence (\eg, the top right corner of $v$ corresponds to the top right corner of $v^\prime$) no longer exists because we apply different spatial augmentations to the two views.
To retrieve the correspondence, we propose two alignment methods: {\it RoI alignment} and {\it offset alignment}.

%-------------------------------------------------------------

\medskip\noindent {\bf (1) RoI alignment.}
Since the IoU threshold guarantees a non-trivial overlapping between $v$ and $v^\prime$, 
{\it RoI alignment} treats the overlapping area as the Region of Interest for both views and uses RoI Align \cite{he2017mask} to extract the feature of intersection region only.
We denote the {\it relative} box coordinates of the overlapping in two views as $B,B^\prime \in \mathcal{R}^4$, then the aligned feature $R,R^\prime \in \mathcal{R}^{H\times W\times C}$ can be represented as:
\vspace{-5mm}
\begin{gather}
    R = RoIAlign(g_{\theta}(F), B), \\
    R^\prime = RoIAlign(g_{\xi}(F^\prime), B^\prime),
\end{gather}
where $g_{\theta}(\cdot)$ and $g_{\xi}(\cdot)$ are the projectors of online and target networks.
If not specified, we keep the spatial resolution of $R$ and $R^\prime$ the same as $F$ and $F^\prime$ by default.

%-------------------------------------------------------------

\medskip\noindent {\bf (2) Offset alignment.}
Although guaranteeing precise feature alignment, {\it RoI alignment} does not make full use of the information of non-overlapping areas.
Motivated by Liu~\etal~\cite{liu2018intriguing}, we provide the coordinate offset map $\Delta C \in \mathcal{R}^{H \times W \times 2}$ from the projected online 2D feature map $G = g_{\theta}(F)$ to the projected target feature map $G^\prime = g_{\xi}(F^\prime)$ as additional information to the predictor for implicit feature alignment, as shown in Figure \ref{fig:misalignment}(c).
Specifically, for $\forall i \in [1,H], j \in [1,W]$, we define the offset map $\Delta C$ as:
\begin{equation}
    \label{equ:offset}\Delta C_{i,j} \triangleq \frac{coord(G^\prime_{i,j}) - coord(G_{i,j})}{coord(G_{H,W}) - coord(G_{1,1})},
\end{equation}
where $coord(\cdot)$ returns the corresponding coordinates of the given pixel in the original image. 
Note that $\Delta C$ should be normalized by the size of $v$ to decrease variance because the predictor with {\it offset alignment} is actually estimating a {\it conditional expectation} (see more details in Section \ref{sec:ablation} and Appendix \ref{sec:proof}). % A).
Then $R$ and $R^\prime$ can be denoted as:
\vspace{-1mm}
\begin{gather}
    R = concat(g_{\theta}(F), \Delta C), \\
    R^\prime = g_{\xi}(F^\prime).
\end{gather}

\vspace{-1mm}
After feature alignment, the model can recover the one-to-one correspondence between $R_{i,j}$ and $R^\prime_{i,j}$ for $\forall i \in [1,H], j \in [1,W]$.
% Based on that, we discuss how to measure the similarity between two 2D inputs in Section \ref{sec:cluster}.

%-------------------------------------------------------------

\subsection{Similarity Measurement of 2D Feature Maps} \label{sec:cluster}

\begin{figure}[t]
\begin{center}
  \includegraphics[width=1.0\linewidth]{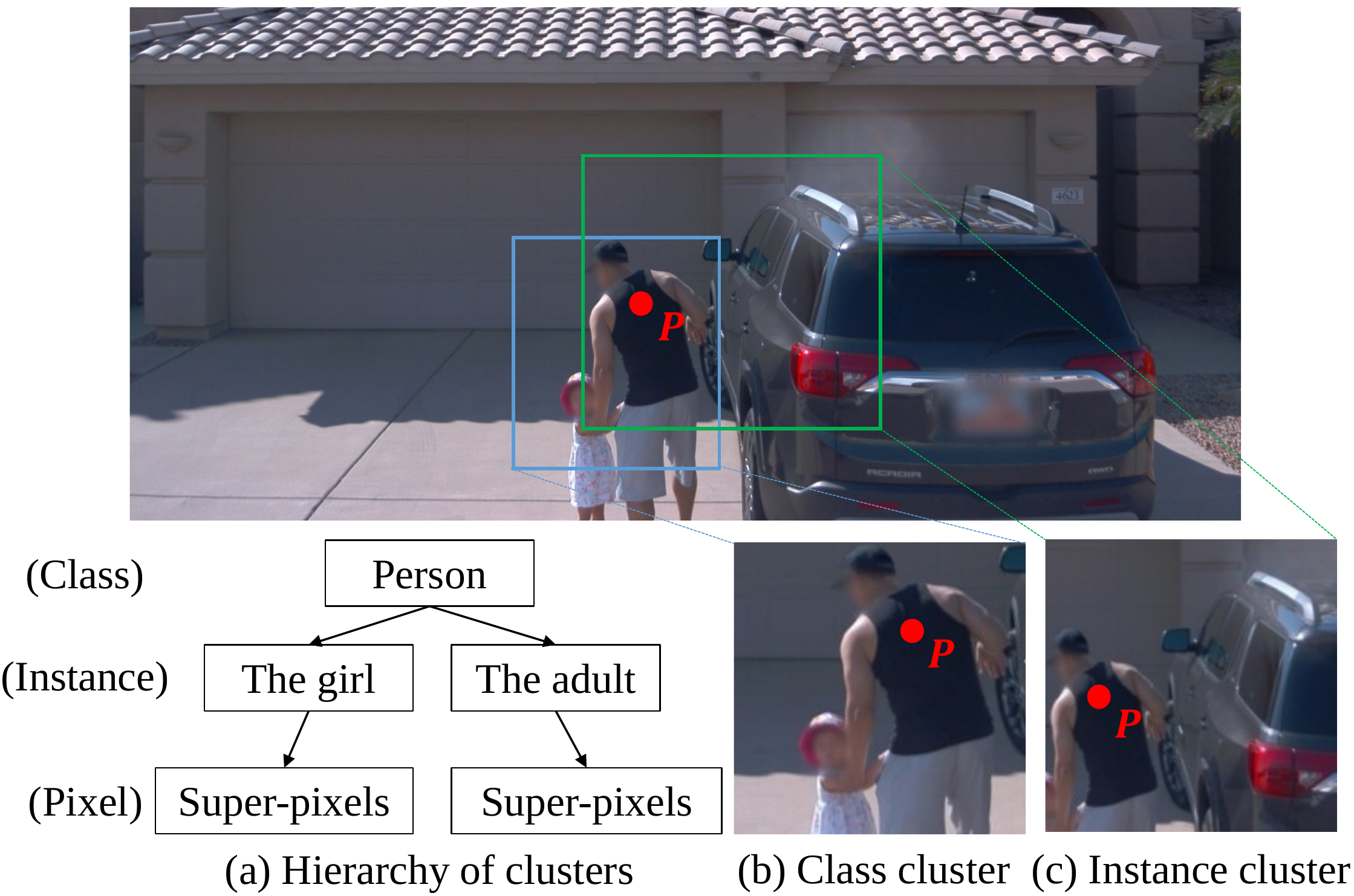}
\end{center}
\caption{{\bf The hierarchy of clusters} on Waymo
(a) from class cluster down to instance and pixel clusters;
(b) {\it class cluster}: pixels belonging to different instances of the same semantic class;
(c) {\it instance cluster}: pixels belonging to the same instance.
Cluster is actually a relative concept.
The same pixel $P$ might belong to different clusters in different contexts of (b) and (c).}
\label{fig:cluster}
\end{figure}

%-------------------------------------------------------------

With 2D feature maps, we can model the relationships between different instances.
We find there exists a hierarchy of clusters in multi-instance circumstances naturally from {\it class cluster} down to {\it instance cluster} and {\it pixel cluster}, as shown in Figure \ref{fig:cluster}(a).
{\it Instance cluster} means that pixels belonging to the same instance should lie in the same cluster, which is consistent with the basic assumption of instance discrimination. On the other hand, {\it class cluster} suggests pixels belonging to different instances of the same semantic class should also form a separate cluster.
Besides, {\it pixel cluster} refers to groups of pixels sharing common characteristics like pixel intensity (\eg, super-pixels).
This hierarchy is more common in autonomous driving datasets (\eg, Waymo \cite{sun2020scalability}) whose semantic class set is relatively small.

%-------------------------------------------------------------

\vspace{-3mm}
\paragraph{Intra-image clustering.} 
Clustering is previously used in SSL mainly for mining {\it inter-image similarity} \cite{caron2018deep,caron2020unsupervised} to reduce false negative during contrastive learning.
However, based on the analysis above, we think clustering can also be performed within a single image to capture {\it intra-image similarity} further.

Specifically, we deploy the K-means algorithm on the aligned target feature map $R^\prime$ to get the assigned cluster {\bf centroid} for each pixel, denoted as $Kmeans(R^\prime)$.
The aligned online feature map $R$ is then fed into the predictor to predict the cluster assignment $q_{\theta}(R)$, which should be consistent with $Kmeans(R^\prime)$ for every pixel due to the one-to-one correspondence after feature alignment.
The network will extract smoother features for pixels in the same cluster so that translation-invariance is also encouraged.

%-------------------------------------------------------------
% Main transfer experiments table

\begin{table*}[t]
\begin{center}
\begin{tabular}{l|c|c|ccc|ccc}
\hline
\multirow{2}*{Method} & \multirow{2}*{Pre-train Dataset} & \multirow{2}*{Epochs} & \multicolumn{3}{c|}{Cityscapes} & \multicolumn{3}{c}{BDD100K} \\
\cline{4-9}
 & & & mAP & AP$_{50}$ & mIoU & mAP & AP$_{50}$ & mIoU \\
\hline %%%%%%%%%%%%%%%%%%%%%%%%%%%%%%
Random Init & - & - & 25.4 & 51.1 & 65.3 & 16.4 & 30.4 & 50.7 \\
Supervised & ImageNet & 90 & 32.9 & 59.6 & 74.6 & 21.9 & 40.0 & 58.8 \\
\hline %%%%%%%%%%%%%%%%%%%%%%%%%%%%%%
InstDist$^\dag$~\cite{wu2018unsupervised} & ImageNet & 200 & 33.0 & 60.1 & 73.3 & 21.4 & 38.9 & 57.2 \\
SwAV$^\dag$~\cite{caron2020unsupervised} & ImageNet & 200 & 33.9 & 62.4 & 73.0 & 22.5 & 40.8 & 57.1 \\
BYOL$^\dag$~\cite{grill2020bootstrap} & ImageNet & 200 & 33.8 & 62.9 & 75.1 & 21.8 & 39.3 & 59.1 \\
MoCo$^\dag$~\cite{he2020momentum} & ImageNet & 200 & 32.3 & 59.3 & 75.3 & 22.4 & 40.4 & 59.7 \\
MoCo-v2$^\dag$~\cite{chen2020improved} & ImageNet & 200 & 33.9 & 60.8 & 75.7 & 23.1 & 41.3 & 60.0 \\
\hline %%%%%%%%%%%%%%%%%%%%%%%%%%%%%%
BYOL~\cite{grill2020bootstrap} & Waymo & 325 & 28.8 & 55.7 & 69.4 & 18.1 & 33.8 & 53.7 \\
MoCo~\cite{he2020momentum} & Waymo & 325 & 30.5 & 57.1 & 73.9 & 21.0 & 39.1 & 57.0 \\
MoCo-v2~\cite{chen2020improved} & Waymo & 325 & 31.4 & 59.4 & 73.6 & 20.9 & 38.9 & 56.6 \\
% \hline %%%%%%%%%%%%%%%%%%%%%%%%%%%%%%
MultiSiam & Waymo & 325 & 31.8$^{+3.0}$ & 59.6$^{+3.9}$ & 74.1$^{+4.7}$ & 21.1$^{+3.0}$ & 39.3$^{+5.5}$ & 57.6$^{+3.9}$ \\
MultiSiam$^{\dag\dag}$ & Waymo & 325 & 32.2$^{+3.4}$ & 59.9$^{+4.2}$ & 75.5$^{+6.1}$ & 21.8$^{+3.7}$ & 40.2$^{+6.4}$ & 56.9$^{+3.2}$ \\
MultiSiam & SODA5M & 55 & {\bf 34.1$^{+5.3}$} & {\bf 61.7$^{+6.0}$} & {\bf 75.8$^{+6.4}$} & {\bf 23.5$^{+5.4}$} & {\bf 42.7$^{+8.9}$} & {\bf 60.3$^{+6.6}$}\\
\hline
\end{tabular}
\end{center}
\vspace{-2mm}
\caption{{\bf Comparisons on Cityscapes and BDD100K instance and semantic segmentation.}
The metrics include mask mAP and AP50 for instance segmentation and mIoU for semantic segmentation.
(1) BYOL, MoCo and MoCo-v2 all suffer from global-inconsistency and performance degradation from ImageNet to Waymo,
(2) but {\it MultiSiam} directly based on BYOL recovers its decrease and shows better generalization ability,
achieving state-of-the-art performance among Waymo pre-trainings.
(3) By pre-training on SODA5M, {\it MultiSiam} exceeds the ImageNet pre-trained MoCo-v2, revealing the potential of domain-specific pre-training.
$^\dag$: we take the officially released pre-trained weights and report fine-tuning results.
$^{\dag\dag}$: a simple implementation of MoCo-based {\it MultiSiam}.
}
\label{tab:transfer}
\vspace{-4mm}
\end{table*}

%-------------------------------------------------------------

\paragraph{Self-attention.}
As shown in Figure \ref{fig:cluster} (b) and (c), cluster is actually a relative concept.
The same pixel $P$ might belong to different clusters under different contexts.
Here we consider the {\it Person} class cluster and the {\it Adult} instance cluster as different clusters since they have different centroids even if ideally {\it Adult} might be a subset of {\it Person}.
However, the predictor $q_{\theta}(\cdot)$ only consists of  $1 \times 1$ convolutions which operate locally.
A ``global" view is needed to deal with the ambiguity of cluster assignments.
Here we deploy the same non-local-network style \cite{wang2018non} self-attention module as Xie~\etal~\cite{xie2020propagate}.
Specifically, for $\forall i \in [1,H], j \in [1,W]$, the final cluster prediction $Q_{i,j}$ of $R_{i,j}$ is defined as:
\begin{equation}
    Q_{i,j} = \sum_{i^\prime=1}^H\sum_{j^\prime=1}^W sim(R_{i,j},R_{i^\prime, j^\prime}) \cdot q_{\theta}(R_{i^\prime,j^\prime}),
\end{equation}
where $q_{\theta}(\cdot)$ is the original local predictor and $sim(\cdot, \cdot)$ is the similarity function defined as:  
\begin{equation}
    sim(R_{i,j},R_{i^\prime, j^\prime}) = (max(cos(R_{i,j},R_{i^\prime, j^\prime}), 0))^2.
\end{equation}
The final 2D clustering consistency loss is defined as:
\begin{equation}
    \mathcal{L}_{2D\_cluster} \triangleq \frac{1}{HW}\sum_{i=1}^H\sum_{j=1}^W -cos(Q_{i,j}, Kmeans(R^\prime_{i,j})).
    % \mathcal{L}_{2D\_cluster} \triangleq \frac{\sum_{i=1}^H\sum_{j=1}^W -cos(Q_{i,j}, Kmeans(R^\prime_{i,j}))}{HW}.
\end{equation}

%-------------------------------------------------------------

\vspace{-3mm}
\paragraph{Incorporated with 1D image-level consistency.}
We also keep the 1D image-level consistency branch as defined in Section \ref{sec:BYOL} due to its effectiveness to improve classification performance.
Both classification and localization are important for visual perception tasks like semantic segmentation.
We formulate {\it MultiSiam} in the multi-task learning manner and the final loss function is a weighted sum of 1D image-level consistency and 2D clustering consistency as:
\begin{equation}
    \mathcal{L}_{MultiSiam} = \lambda \mathcal{L}_{1D\_img} + (1 - \lambda) \mathcal{L}_{2D\_cluster},
\end{equation}
where the balanced weight $\lambda$ is set to be 0.5 by default.

%-------------------------------------------------------------

\section{Experiments}

\subsection{Implementation Details}

\paragraph{Dataset.} 
We pre-train our {\it MultiSiam} on the widely used Waymo Open~\cite{sun2020scalability} autonomous driving dataset mainly, which consists of around 790 thousand training images.
The image sizes range from (1920, 968) to (1920, 1280).

\vspace{-4mm}
\paragraph{Data augmentation.}
We follow the standard data augmentation pipeline in BYOL \cite{grill2020bootstrap} with the proposed IoU threshold. 
Two random crops whose IoU is larger than the pre-defined threshold (0.5 by default) are generated and then resized to $224 \times 224$, followed by random horizontal flip, color jitter, Gaussian blur and solarization.
We flip back the projected 2D feature map before feature alignment if the horizontal flip is applied previously.
All the augmentation parameters are kept the same with BYOL.

\vspace{-4mm}
\paragraph{Training details.}
We adopt standard ResNet-50 \cite{he2016deep} as the backbone network.
The momentum starts at 0.996 and increases gradually to 1.0 at the end of the training process.
We use the LARS \cite{you2017large} optimizer with a cosine learning rate scheduler for large batch training.
The base learning rate is set to 1.0, which will scale linearly with the batch size ($lr = lr_{base} \times bs / 256$).
Weight decay is set to 1e-5.
We use a batch size of 1024 running on 8 Tesla V100 GPUs per experiment.
We pre-train 325 epochs on Waymo to maintain similar training iterations with training 200 epochs on ImageNet for a fair comparison. 
For ablation studies, we adopt 150 epochs pre-training and then evaluate on Cityscapes~\cite{cordts2016cityscapes} \texttt{val} set for semantic segmentation.

%-------------------------------------------------------------

\subsection{Transfer Settings and Results}\label{sec:transfer}

\paragraph{Baseline.}
In this paper, we choose the strong BYOL \cite{grill2020bootstrap}, MoCo \cite{he2020momentum} and MoCo-v2 \cite{chen2020improved} as our baseline methods.
We use OpenSelfSup~\footnote{https://github.com/open-mmlab/OpenSelfSup} as our codebase to pre-train all baseline methods on Waymo and
tune the hyperparameters according to the transfer results on Cityscapes \texttt{val} set of 150 epochs pre-training for better performance.

\vspace{-4mm}
\paragraph{Transfer settings.} 
We choose widely used Cityscapes~\cite{cordts2016cityscapes} and BDD100K~\cite{yu2018bdd100k} instance and semantic segmentation for autonomous driving as downstream tasks, adapting the settings of Cityscapes in MoCo~\cite{he2020momentum} for both datasets using Detectron2~\cite{wu2019detectron2}.
For instance segmentation, we fine-tune a Mask R-CNN detector (FPN-backbone) for 24k iterations, while a FCN-16s~\cite{long2015fully} is trained for 90k iterations for semantic segmentation.
We train the models on both \texttt{train} sets and evaluate on the corresponding \texttt{val} sets separately.

%-------------------------------------------------------------

%%%%%%%%%%%%%%%%%%%%%%%%%%%%%%%%%%%%%%%
\begin{table}
\begin{center}
\begin{tabular}{c|c|c|c|c}
\hline
\multirow{2}*{IoU Thre} & \multicolumn{3}{{c|}}{Feature Alignment} & \multirow{2}*{mIoU} \\
\cline{2-4}
 & RoI & Offset & Offset nonorm & \\
\hline
\multicolumn{5}{{l}}{(a) IoU threshold} \\
\hline
0.3 & \checkmark & & & 69.2 \\
0.4 & \checkmark & & & 68.6 \\
0.5 & \checkmark & & & {\bf 70.0} \\
0.6 & \checkmark & & & 69.7 \\
0.7 & \checkmark & & & 69.2 \\
\hline
\multicolumn{5}{{l}}{(b) Feature alignment methods} \\
\hline
0.5  &  & & & 68.8 \\
0.5 & \checkmark & & & {\bf 70.0} \\
0.5 &  & \checkmark & & 69.5 \\
0.5 &  &  & \checkmark & 69.4 \\
\hline
\end{tabular}
\end{center}
\vspace{-2mm}
\caption{{\bf Ablations on positive sample definition.} 
(a) IoU threshold; (b) feature alignment. 
All results are evaluated on Cityscapes \texttt{val} set over three independent trials.}
\label{tab:IO}
\vspace{-4mm}
\end{table}

%%%%%%%%%%%%%%%%%%%%%%%%%%%%%%%%%%%%%%%

%-------------------------------------------------------------

\vspace{-4mm}
\paragraph{Discussion.}
We report the final transfer results in Table~\ref{tab:transfer}.
All baseline methods suffer from significant performance degradation adapted from ImageNet to Waymo, revealing the vulnerability to global-inconsistency of current models.
Our {\it MultiSiam}, however, recovers the decrease and generalizes better to multi-instance circumstances.
Compared with our direct BYOL baseline, {\it MultiSiam} improves {\bf 3.0\%/3.0\%} mAP and {\bf 4.7\%/3.9\%} mIoU for instance and semantic segmentation separately, 
surpassing the strong MoCo baselines and achieving state-of-the-art performance among Waymo pre-trainings.
To verify the guidance effect of clustering, we also train {\it MultiSiam without K-means} (more details in Appendix \ref{sec:wo_kmeans}) % B.1) 
and achieve 71.0\% and 54.8\% mIoU on Cityscapes and BDD100K, which is significantly worse than {\it MultiSiam}, suggesting that K-means might help produce more robust targets during self-supervised learning.
% Note that MoCo transfers better than MoCo-v2, which suggests the MLP in MoCo-v2 is also related to single-centric-object and intra-image similarity mining.

\vspace{-3mm}
\paragraph{Domain-specific pre-training.}
Although being a self-driving dataset, Waymo still has disadvantages in both quantity (0.79M vs 1.28M) and quality (\eg, imbalance of foreground and background) compared with ImageNet, which may hurt the performance of Waymo representations.
We further pre-train {\it MultiSiam} on SODA10M~\cite{han2021soda10m}, a large-scale autonomous driving dataset, with similar GPU hours of ImageNet pre-trainings and surpass the ImageNet pre-trained MoCo-v2 as shown in Table~\ref{tab:transfer}, revealing the potential of domain-specific pre-training.
Due to the hardware resources, here we only use a 5-million subset of SODA10M (split 0, 2, 4, 6, 8), denoted as SODA5M.
Note that due to single-centric-object guarantee, it is costly to collect a large amount of ImageNet-like samples, making multi-instance self-supervised learning more valuable in practice.

% What is more, the single-centric-object guarantee can be considered as extra processing cost of ImageNet. 
% With {\it MultiSiam} to deal with the multi-instance data, it is possible to perform self-supervised pre-training in a larger scale of street scene datasets in future work.

\vspace{-4mm}
\paragraph{Flexibility.}
As shown in Table \ref{tab:transfer}, BYOL has a significant performance gap with MoCo because pure cross-view consistency framework might be more vulnerable to global-inconsistency than contrastive methods, which is complementary to the improvement of {\it MultiSiam}.
As a flexible plug-and-play module, {\it MultiSiam} can be naturally extended to contrastive learning based SSL methods.
Here we deploy a simple implementation of MoCo-based {\it MultiSiam} (more details in Appendix \ref{sec:moco_based}) % B.2) 
and obtain further improvements (\eg, 0.4\% mAP and 1.4\% mIoU on Cityscapes in Table~\ref{tab:transfer}).
More delicate design is definitely needed to achieve the best performance, which we will dig into in future work.

% As a plug-and-play module, 
% We will try to incorporate contrastive learning into {\it MultiSiam} and perform self-supervised pre-training in a larger scale of street scene datasets in our future work.

\vspace{-4mm}
\paragraph{Generalization.}
Although originally designed for multi-instance circumstances, our {\it MultiSiam} also demonstrates remarkable generalization ability to single-centric-object datasets. As shown in Appendix \ref{sec:IN}, % C, 
ImageNet-pretrained {\it MultiSiam} still achieves state-of-the-art performance even when compared with more recent SSL methods~\cite{wang2020dense,xie2021detco}.

%-------------------------------------------------------------

\subsection{Ablation Study and Analysis}\label{sec:ablation}

%%%%%%%%%%%%%%%%%%%%%%%%%%%%%%%%%%%%%%%

\begin{table}
\setlength{\tabcolsep}{3.5mm}
\begin{center}
\begin{tabular}{c|c|c|c}
\hline
\#Cluster K & Dist metrics & Dense & mIoU \\
\hline
\multicolumn{4}{{l}}{(a) Number of Clusters K} \\
\hline
3 & Cosine &  & {\bf 70.0} \\
4 & Cosine &  & 68.4 \\
5 & Cosine &  & 69.7 \\
\hline
\multicolumn{4}{{l}}{(b) Dense Clustering \& Distance Metrics} \\
\hline
3 & Cosine &  & {\bf 70.0} \\
3 & Cosine & \checkmark & 69.0 \\
3 & Euclidean &  & 69.3 \\
\hline
% \multicolumn{4}{{l}}{(c) } \\
% \hline
% 3 & Cosine &  & {\bf 70.0} \\
% \hline
\end{tabular}
\end{center}
\vspace{-2mm}
\caption{{\bf Ablations on intra-image clustering.} 
(a) Cluster number K; (b) dense clustering and clustering distance metrics.}
\vspace{-2mm}
\label{tab:Kmeans}
\end{table}

%%%%%%%%%%%%%%%%%%%%%%%%%%%%%%%%%%%%%%%%%%%%%%%

%%%%%%%%%%%%%%%%%%%%%%%%%%%%%%%%%%%%%%%
\begin{table}
\begin{center}
\begin{tabular}{c|c|c|c}
\hline
Method & Components & mIoU & $\Delta$ \\
\hline
BYOL &  & 67.2 &  \\
\hline
\multirow{4}*{MultiSiam} & + IoU thre 0.5 & 69.2 & + 2.0 \\
& + $\mathcal{L}_{cluster}$ & 70.0 & + 0.8 \\
& + self-attention & 71.2 & + 1.2 \\
& + offset align & {\bf 71.9} & + 0.7 \\
& /+ offset align nonorm & 71.2 & + 0.0 \\
\hline
\end{tabular}
\end{center}
\vspace{-2mm}
\caption{{\bf Ablations on proposed components.} 
The model's performance continues to increase after each component is deployed.}
\label{tab:variant}
\vspace{-4mm}
\end{table}

%%%%%%%%%%%%%%%%%%%%%%%%%%%%%%%%%%%%%%%%%%%%%%%%%%

%-------------------------------------------------------------

\paragraph{Setup.}
We conduct 150 epochs pre-training with a base learning of 0.3 for all ablation studies.
The batch size is set to 1024, and parameters are updated every four iterations to mimic a batch size of 4096 following the BYOL \cite{grill2020bootstrap}.
We report the average result of three independent trials to reduce variance.
See more ablations in Appendix \ref{sec:more_ablation}. % D.

%-------------------------------------------------------------

\vspace{-4mm}
\paragraph{IoU threshold.}
As shown in Table \ref{tab:variant}, BYOL simply equipped with the proposed IoU threshold during random cropping can achieve a significant improvement of 2.0\% mIoU, showing the effectiveness of local-consistency.
We further find that there is a balanced point when the IoU threshold is set to 0.5 in Table \ref{tab:IO}(a).
Interestingly, the model achieves sup-optimal results when the threshold is set to 0.7, which can generate well-aligned random crops in Figure~\ref{fig:IoU}(d).
We think it is because by controlling how ``close" the two crops are, the IoU threshold also controls a trade-off between {\bf noise} and {\bf data complexity}.
Remember that the motivation of data augmentation is to generate two {\bf different} but {\bf consistent} views.
We do not want to differ the two views too much since it will violate the basic assumption of instance discrimination. 
Meanwhile, the two views should not be exactly the same. Otherwise, the network would tend to collapse due to the trivial supervision signal.

%-------------------------------------------------------------

\vspace{-4mm}
\paragraph{Feature alignment.}
We ablate the proposed two feature alignment methods in Table \ref{tab:IO}(b).
The model suffers from a performance degradation of 1.2\% mIoU without feature alignment due to the absence of one-to-one correspondence.
Both {\it RoI} and {\it offset alignment} achieve significant improvements and as we can see in Table \ref{tab:variant}, {\it offset alignment} performs much better when self-attention is available.

We also verify the necessity of coordinate normalization during the offset alignment in Equation (\ref{equ:offset}).
As shown in Table~\ref{tab:IO}(b) and \ref{tab:variant}, offset alignment with coordinate normalization performs better whether or not self-attention is available. 
We analyze effectiveness of offset alignment based on the EM-like hypothesis proposed by Chen~\etal~\cite{chen2020exploring}.
Specifically, predictor with offset alignment is actually estimating an expectation of target cluster centroids conditioned on the online feature and offset map (see proofs in Appendix \ref{sec:proof}): % A):
\vspace{-2mm}
\begin{equation}
    \label{equ:exp}q^{optimal}_{\theta}(R, \Delta C) = \mathbb{E}\big[Kmeans(R^\prime) | \Delta C, R)\big].
\end{equation}
So to reduce variance during training, using normalized offset maps should be a better choice than the absolute values.

%-------------------------------------------------------------

\vspace{-4mm}
\paragraph{Number of clusters K.}
We ablate the number of clusters K during the K-means algorithm in Table \ref{tab:Kmeans}(a).
The optimal performance is obtained when $K = 3$.
One of the reasons is that self-driving datasets like Waymo usually have a relatively small semantic class set (\eg size 4 for Waymo).
Besides, as we discuss in Section \ref{sec:cluster},
the definition of clusters in multi-instance circumstances is actually a relative concept dependent on the specific context of a random crop, which is not limited to {\it instance} or {\it class} only.
% For example, even the clusters of "foreground" and "background" are possible in multi-instance images.
Keeping the number of clusters $K$ at a relatively small value will make the model more robust to different image contexts.

%-------------------------------------------------------------

\vspace{-4mm}
\paragraph{Dense clustering.}
Instead of predicting cluster centroids, dense clustering calculates cosine similarity with every pixel in the corresponding cluster.
As shown in Table~\ref{tab:Kmeans}(b), predicting centroids performs better because as discussed in Equation (\ref{equ:exp}), the predictor estimates a conditional expectation while dense clustering will increase variance.

%-------------------------------------------------------------

\vspace{-4mm}
\paragraph{Self-attention.}
As we can see in Table \ref{tab:variant}, using the local predictor can only achieve a marginal improvement of 0.8\% mIoU upon BYOL with 0.5 IoU threshold because of the ambiguity of cluster assignments in Section~\ref{sec:cluster}.
After introducing a ``global" view by deploying self-attention with the predictor, our model can get more precise clustering predictions and achieve another significant improvement.
Note that offset alignment performs much better together with self-attention, because introducing relative offset can make better use of the image locality for clustering prediction.

%------------------------------------------------------------

\vspace{-1mm}
\subsection{Visualization}
\vspace{-1mm}

\begin{figure}[t]
\begin{center}
  \includegraphics[width=0.76\linewidth]{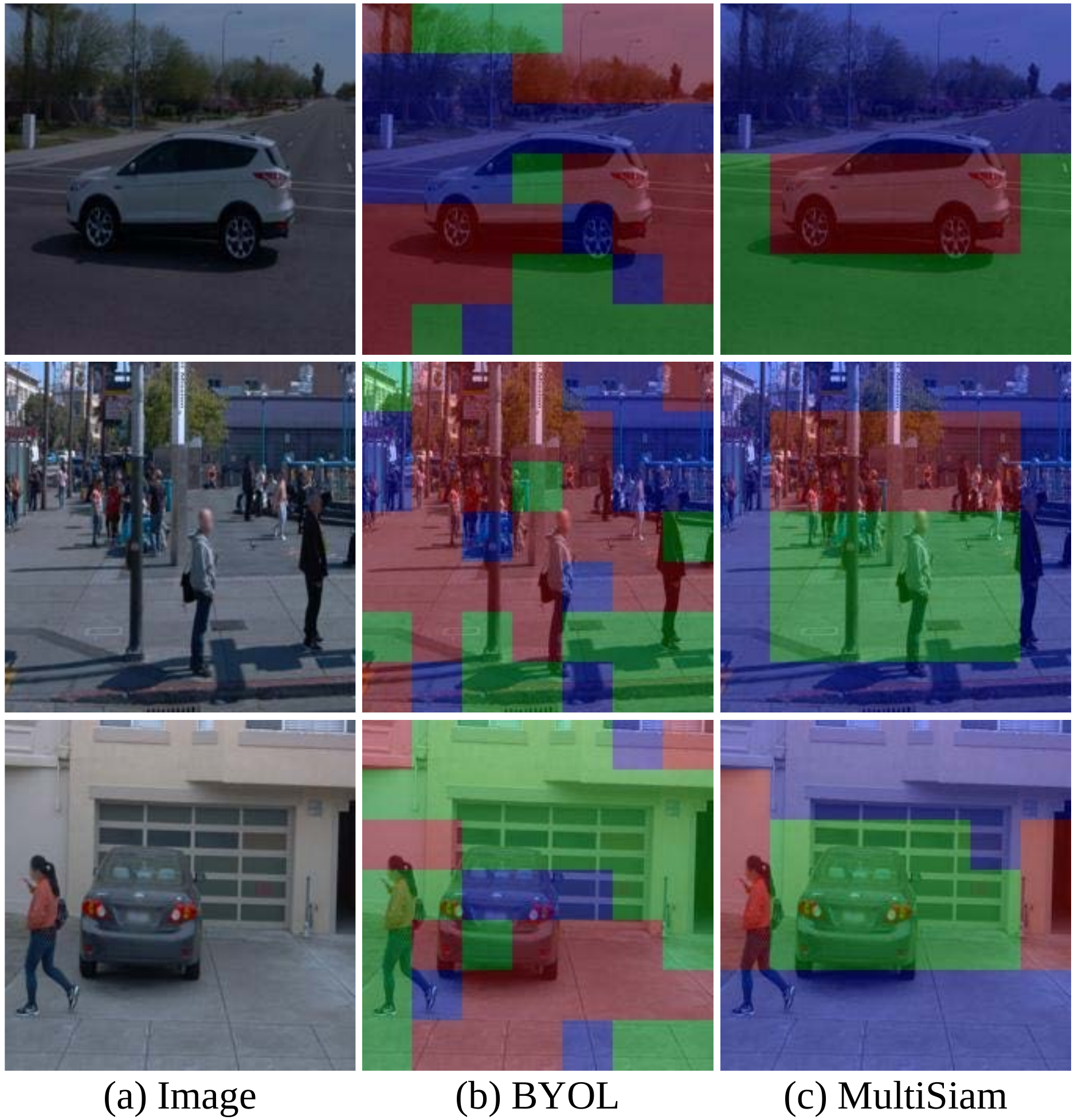}
\end{center}
\vspace{-3mm}
\caption{\textbf{Clustering ($K = 3$) results of BYOL and \textit{MultiSiam}.} Different colors represent different clusters.
{\it MultiSiam} can better capture intra-image similarity in different circumstances.
}
\label{fig:visualization}
\vspace{-5mm}
\end{figure}

%------------------------------------------------------------

Figure~\ref{fig:visualization} shows the clustering results on the final backbone 2D feature maps of BYOL and {\it MultiSiam} in single instance, multi-instance of the same class and multi-instance of different classes circumstances.
BYOL shows random clustering pattern while {\it MultiSiam} can better capture intra-similarity.
It recognizes the car in the first row, separates the background crowds from the man in the second row and distinguishes the lady and the car successfully in the third row.
The clustering result of {\it MultiSiam} is much smoother, suggesting that translation-invariance is also enhanced.

%------------------------------------------------------------

\vspace{-1mm}
\section{Conclusion}
\vspace{-1mm}
% This paper explores the usage of SSL in general multi-instance autonomous driving circumstances.
% We show that current methods rely on the single-centric-object ImageNet suffer from performance drop using multi-instance datasets like Waymo for pre-training.
% Targeting this issue, our proposed method, {\it MultiSiam}, demonstrates better generalization ability and achieves state-of-the-art transfer performance by dealing with positive sample definition and adopting intra-image clustering with self-attention.
% We believe the information in multi-instance data is still under-explored (\eg, long distance similarity).
% We hope our simple yet effective method can bring researchers' attention to multi-instance self-supervised learning. 

This paper explores self-supervised learning in multi-instance autonomous driving circumstances.
We show that current methods which rely on the single-centric-object ImageNet suffer from performance degradation on multi-instance datasets like Waymo for pre-training.
Targeting this issue, our proposed {\it MultiSiam} demonstrates better generalization and achieves state-of-the-art transfer performance by dealing with positive sample design and adopting intra-image clustering with self-attention.
We believe the information in multi-instance data is still under-explored (\eg, long distance similarity).
We hope our simple yet effective method can bring researchers' attention to multi-instance self-supervised learning.

%------------------------------------------------------------

\appendix

\section*{Appendix}

%------------------------------------------------------------
%%%%%%%%%%%%%%%%%%%%
% Reference:
% Equation (12) * 3
% Section * 3
% Appendix * 6 (in main body)
%%%%%%%%%%%%%%%%%%%%

% \appendix

% \section*{Appendix}

\section{Proof for Equation (\ref{equ:exp})} % (12)
\label{sec:proof}

We prove Equation (\ref{equ:exp}) % (12)
similarly with Chen~\etal~\cite{chen2020exploring}.
All the terminologies are kept consistent with Section \ref{sec:method}. % 3.
We first consider the case when momentum $\tau=0$ (\ie, the parameters of online and target networks are always the same).
By definition, for $\forall i \in [1,H], j \in [1,W]$, the predictor $q_\theta(\cdot)$ is expected to minimize:
\begin{equation}
    \mathbb{E}_r\bigg[||q_\theta(R_{i,j}, \Delta C_{i,j}) - Kmeans(R^\prime_{i,j}) ||_2^2 \bigg],
\end{equation}
where $r$ is a random variable representing the feature of a random view (\eg, data augmentation) of image $x$.
For the simplicity of analysis, we use the mean square root $||\cdot||_2^2$ here, which is equivalent to cosine distance after the vectors are $l_2$-normalized.
Then the optimal solution to $q_\theta(\cdot)$ should satisfy:
\begin{equation}
    \label{equ:exp_pixel}
    q_\theta^{optimal}(R_{i,j}, \Delta C_{i,j}) = \mathbb{E}_r\bigg[Kmeans(R^\prime_{i,j}) | \Delta C_{i,j}, R_{i,j} \bigg],
\end{equation}
for any pixel of any image, which is equivalent to Equation (\ref{equ:exp}). % (12).
Note that, here $r$ denotes a conditional distribution conditioned on the online feature map $R$ and the offset map $\Delta C$, instead of a uniform distribution in Chen~\etal~\cite{chen2020exploring}.
When $\tau\neq 0$, the target network is a exponential moving average of the online network, which also helps estimate the expectation in Equation~(\ref{equ:exp_pixel}), as suggested in Chen~\etal~\cite{chen2020exploring}.

%------------------------------------------------------------

%%%%%%%%%%%%%%%%%%%%%%%%%%%%%%%%%%%%%%%%%%%%%%%
% ImageNet experiments Table

\begin{table*}[t]
\begin{center}
\scalebox{0.93}{
\begin{tabular}{l|ccc|ccc|ccc|ccc|c}
\hline %%%%%%%%%%%%%%%%%%%%%%%%%%%%%%%%%%%%%%%%%%%%%%%
\multirow{2}*{Method} & \multicolumn{3}{c|}{PASCAL VOC} & \multicolumn{6}{c|}{COCO Mask R-CNN 90k (1x)} & \multicolumn{3}{c|}{Cityscapes} & BDD100K \\
\cline{2-14}
 & AP & AP$_{50}$ & AP$_{75}$ & AP$^{bb}$ & AP$^{bb}_{50}$ & AP$^{bb}_{75}$ & AP$^{mk}$ & AP$^{mk}_{50}$ & AP$^{mk}_{75}$ & AP & AP$_{50}$ & mIOU & mIOU \\
\hline %%%%%%%%%%%%%%%%%%%%%%%%%%%%%%%%%%%%%%%%%%%%%%%
Rand Init & 33.8 & 60.2 & 33.1 & 31.0 & 49.5 & 33.2 & 28.5 & 46.8 & 30.4 & 25.4 & 51.1 & 65.3 & 50.7 \\
Supervised & 53.5 & 81.3 & 58.8 & 38.9 & 59.6 & 42.7 & 35.4 & 56.5 & 38.1 & 32.9 & 59.6 & 74.6 & 58.8 \\
\hline %%%%%%%%%%%%%%%%%%%%%%%%%%%%%%%%%%%%%%%%%%%%%%%
InstDist & 55.2 & 80.9 & 61.2 & 37.4 & 57.6 & 40.6 & 34.1 & 54.6 & 36.4 & 33.0 & 60.1 & 73.3 & 57.2 \\
SwAV & 56.1 & 82.6 & 62.7 & 38.5 & 60.4 & 41.4 & 35.4 & 57.0 & 37.7 & 33.9 & 62.4 & 73.0 & 57.1 \\
MoCo & 55.9 & 81.5 & 62.6 & 38.5 & 58.9 & 42.0 & 35.1 & 55.9 & 37.7 & 32.3 & 59.3 & 75.3 & 59.7 \\
MoCo-v2 & 57.0 & 82.4 & 63.6 & 38.9 & 59.4 & 42.4 & 35.5 & 56.5 & 38.1 & 33.9 & 60.8 & 75.7 & 60.0 \\
\hline %%%%%%%%%%%%%%%%%%%%%%%%%%%%%%%%%%%%%%%%%%%%%%%
DetCo & 57.8 & 82.6 & 64.2 & 39.5 & 60.3 & 43.1 & 35.9 & 56.9 & 38.6 & 34.7 & 63.2 & 76.5 & 60.9 \\
DenseCL & 58.7 & 82.8 & 65.2 & 40.3 & 59.9 & 44.3 & 36.4 & 57.0 & 39.2 & 34.3 & 62.5 & 75.7 & 59.3 \\
\hline %%%%%%%%%%%%%%%%%%%%%%%%%%%%%%%%%%%%%%%%%%%%%%%
MultiSiam & 57.8 & \textbf{83.0} & 65.0 & \textbf{40.7} & \textbf{61.7} & \textbf{44.5} & \textbf{37.0} & \textbf{58.6} & \textbf{39.7} & \textbf{34.9} & \textbf{63.8} & \textbf{77.2} & \textbf{61.7}  \\
\hline
\end{tabular}}
\end{center}
\caption{{\bf Comparisons between methods pre-trained on ImageNet.}
Although designed for multi-instance circumstances, our proposed {\it MultiSiam} still achieves state-of-the-art performance and demonstrates strong generalization ability to single-centric-object datasets.
}
\vspace{-2mm}
\label{tab:imagenet}
\end{table*}

%%%%%%%%%%%%%%%%%%%%%%%%%%%%%%%%%%%%%%%%%%%%%%%
% 1st more results on COCO

\begin{table*}[ht]
\setlength\tabcolsep{5pt}
\begin{center}
\scalebox{0.93}{
\begin{tabular}{l|ccc|ccc|ccc|ccc|ccc}
\hline
\multirow{2}*{Method} & \multicolumn{6}{c|}{Mask R-CNN 180k (2x)} & \multicolumn{3}{c|}{RetinaNet 90k (1x)} & \multicolumn{3}{c|}{RetinaNet 180k (2x)} & \multicolumn{3}{c}{RetinaNet 12k} \\
\cline{2-16}
 & AP$^{bb}$ & AP$^{bb}_{50}$ & AP$^{bb}_{75}$ & AP$^{mk}$ & AP$^{mk}_{50}$ & AP$^{mk}_{75}$ & AP & AP$_{50}$ & AP$_{75}$ & AP & AP$_{50}$ & AP$_{75}$ & AP & AP$_{50}$ & AP$_{75}$ \\
\hline %%%%%%%%%%%%%%%%%%%%%%%%%%%%%%%%%%%%%%%%%%%%%%%
Rand Init & 36.7 & 56.7 & 40.0 & 33.7 & 53.8 & 35.9 & 24.5 & 39.0 & 25.7 & 32.2 & 49.4 & 34.2 & 4.0 & 7.9 & 3.5 \\
Supervised & 40.6 & 61.3 & 44.4 & 36.8 & 58.1 & 39.5 & 37.4 & 56.5 & 39.7 & 38.9 & 58.5 & 41.5 & 24.3 & 40.7 & 25.1 \\
\hline %%%%%%%%%%%%%%%%%%%%%%%%%%%%%%%%%%%%%%%%%%%%%%%
MoCo & 40.8 & 61.6 & 44.7 & 36.9 & 58.4 & 39.7 & 36.3 & 55.0 & 39.0 & 38.7 & 57.9 & 41.5 & 20.2 & 33.9 & 20.8 \\
MoCo-v2 & 40.9 & 61.5 & 44.7 & 37.0 & 58.7 & 39.8 & 37.2 & 56.2 & 39.6 & 39.3 & 58.9 & 42.1 & 22.2 & 36.9 & 23.0 \\
DetCo & 41.5 & 62.1 & 45.6 & 37.6 & 59.2 & 40.5 & 38.0 & 57.4 & 40.7 & 39.8 & 59.5 & 42.4 & 23.6 & 38.7 & 24.6 \\
\hline
MultiSiam & \textbf{42.1} & \textbf{63.2} & \textbf{46.1} & \textbf{38.2} & \textbf{60.2} & \textbf{41.1} & \textbf{38.4} & \textbf{57.9} & \textbf{41.2} & \textbf{40.0} & \textbf{59.6} & \textbf{42.8} & \textbf{23.8} & \textbf{39.8} & 24.5\\
\hline
\end{tabular}}
\end{center}
\caption{{\bf Comparisons on COCO objection detection and instance segmentation.}
All methods are pre-trained on ImageNet for 200 epochs.
As we can see, {\it MultiSiam} outperforms all self-supervised counterparts in all downstream settings.
}
\vspace{-2mm}
\label{tab:coco}
\end{table*}

%%%%%%%%%%%%%%%%%%%%%%%%%%%%%%%%%%%%%%%%%%%%%%%
% 2nd more results on COCO

\begin{table}[ht]
\setlength\tabcolsep{5pt}
\begin{center}
\scalebox{0.93}{
\begin{tabular}{l|ccc|ccc}
\hline
\multirow{2}*{Method} & \multicolumn{6}{c}{Mask R-CNN 12k} \\
\cline{2-7}
 & AP$^{bb}$ & AP$^{bb}_{50}$ & AP$^{bb}_{75}$ & AP$^{mk}$ & AP$^{mk}_{50}$ & AP$^{mk}_{75}$  \\
\hline %%%%%%%%%%%%%%%%%%%%%%%%%%%%%%%%%%%%%%%%%%%%%%%
Rand Init & 10.7 & 20.7 & 9.9 & 10.3 & 19.3 & 9.6 \\
Supervised & 28.4 & 48.3 & 29.5 & 26.4 & 45.2 & 25.7 \\
\hline %%%%%%%%%%%%%%%%%%%%%%%%%%%%%%%%%%%%%%%%%%%%%%%
InstDist & 24.2 & 41.5 & 25.1 & 22.8 & 38.9 & 23.7 \\
SwAV & 25.5 & 46.2 & 25.4 & 24.8 & 43.5 & 25.3 \\
MoCo & 25.6 & 43.4 & 26.6 & 23.9 & 40.8 & 24.8 \\
MoCo-v2 & 26.6 & 44.9 & 27.7 & 24.8 & 42.1 & 25.7 \\
DetCo & 27.9 & 46.9 & 29.3 & 26.0 & 44.2 & 26.9 \\
\hline %%%%%%%%%%%%%%%%%%%%%%%%%%%%%%%%%%%%%%%%%%%%%%%
MultiSiam & \textbf{30.3} & \textbf{50.6} & \textbf{31.8} & \textbf{28.5} & \textbf{47.8} & \textbf{29.8} \\
\hline
\end{tabular}}
\end{center}
\caption{{\bf Comparisons on COCO objection detection and instance segmentation by training 12k iterations.}
Our {\it MultiSiam} exceeds all baseline methods with a larger margin compared to fine-tuning 90k and 180k iterations.
}
\vspace{-2mm}
\label{tab:coco_2}
\end{table}

%%%%%%%%%%%%%%%%%%%%%%%%%%%%%%%%%%%%%%%%%%%%%%%

%------------------------------------------------------------

\section{More Implementation Details}\label{sec:more_imple}

\subsection{MultiSiam without K-means}\label{sec:wo_kmeans}

We further implement a {\it MultiSiam without K-means} in Section \ref{sec:transfer} % 4.2 
by calculating a per-pixel cosine distance of 2D features with all other modules unchanged to verify the guidance effect of K-means.
Specifically, the consistency loss between the online network's prediction $Q$ and the aligned target feature $R^\prime$ is now defined as:
\begin{equation}
    \mathcal{L}_{2D\_wo\_cluster} \triangleq \frac{1}{HW}\sum_{i=1}^H\sum_{j=1}^W -cos(Q_{i,j}, R^\prime_{i,j}),
\end{equation}
without adopting K-means clustering on the target network.
All other modules including the IoU threshold, the feature alignment and the self-attention remain unchanged.

%------------------------------------------------------------

\subsection{MoCo-based MultiSiam}\label{sec:moco_based}

Here we briefly introduce our simple implementation of MoCo-based {\it MultiSiam} with {\it RoI alignment}.
All the terminologies are kept consistent with Section \ref{sec:method}. % 3. 
Since MoCo does not adopt a separate predictor, feature alignment is now deployed before projectors instead of after as the BYOL-based {\it MultiSiam} does.
Specifically, the aligned online and target feature maps $R$ and $R^\prime$ are represented as:
\vspace{-3mm}
\begin{gather}
    R = RoIAlign(F, B), \\
    R^\prime = RoIAlign(F^\prime, B^\prime),
\end{gather}
which will be fed into the projectors to get the projected online and target feature maps $G$ and $G^\prime$ as:
\begin{gather}
    G_{i,j} = \sum_{i^\prime=1}^H\sum_{j^\prime=1}^W sim(R_{i,j},R_{i^\prime, j^\prime}) \cdot g_{\theta}(R_{i^\prime,j^\prime}), \\
    G^\prime=g_\xi(R^\prime).
\end{gather}
Note that self-attention module is now incorporated into the online projector.
K-means is then performed on $G^\prime$ to get the centroid of each pixel, denoted as $Kmeans(G^\prime_{i,j})$, which will be considered as the positive sample for the pixel at the same relative position of the online feature map (\ie, $G_{i,j}$).
The final per-pixel contrastive loss $\mathcal{L}_{i,j}$ and the MoCo-based 2D clustering consistency loss $\mathcal{L}_{2D\_cluster}$ is defined as (for $\forall i \in [1,H], j \in [1,W]$):
\begin{gather}
    \mathcal{L}_{2D\_cluster} \triangleq \frac{1}{HW}\sum_{i=1}^H\sum_{j=1}^W\mathcal{L}_{i,j}, \\
    \mathcal{L}_{i,j} \triangleq \frac{exp(G_{i,j} \cdot k^+_{i,j} / \tau)}{exp(G_{i,j} \cdot k^+_{i,j} / \tau) + \Sigma_{k^-} exp(G_{i,j} \cdot k^- / \tau)},
\end{gather}
where $k^+_{i,j} = Kmeans(G^\prime_{i,j})$ is the positive sample of $G_{i,j}$, while $\{k^-\}$ are negative samples maintained by a separate momentum queue containing features from different images following DenseCL~\cite{wang2020dense}.
$\tau$ is the temperature hyper-parameter, which is set to be 0.2 by default.

%------------------------------------------------------------

\section{Pre-training on ImageNet}\label{sec:IN}

\paragraph{Setup.}
We further pre-train our {\it MultiSiam} on the single-centric-object ImageNet dataset to verify its generalization ability. 
All hyper-parameters are kept the same with Waymo pre-trainings without specific tuning.
We pre-train for 200 epochs and report the transfer performance on different downstream tasks.
Note that here we cite the results of DetCo~\cite{xie2021detco} without using  Rand-Augmentation during pre-training for a fair comparison with other methods.

%%%%%%%%%%%%%%%%%%%%%%%%%%%%%%%%%%%%%%%%%%%%%%%

\paragraph{Downstream tasks.}
We choose six representative downstream tasks to evaluate the features following DetCo~\cite{xie2021detco}, including object detection on Pascal VOC~\cite{everingham2010pascal}, objection detection and instance segmentation on COCO~\cite{lin2014microsoft}, instance and semantic segmentation on Cityscapes~\cite{cordts2016cityscapes} and semantic segmentation on BDD100K~\cite{yu2018bdd100k}.

For VOC object detection, we train a Faster R-CNN (C4-backbone) on VOC \texttt{trainval07+12} set for 24k iterations and evaluate on VOC \texttt{test} set.
We evaluate COCO object detection and instance segmentation by fine-tuning on COCO \texttt{train2017} set and test on COCO \texttt{val2017} set.
Here we adopt both two-stage Mask R-CNN (FPN-backbone) and one stage RetinaNet with three training schedules, including standard 90k (1x), 180k (2x) in~\cite{wu2019detectron2} and an extreme 12k-iteration schedule following DetCo for fast convergence, since it is possible to get competitive results on COCO even training from scratch but with enough iterations.
For Cityscapes instance segmentation, we train a Mask R-CNN (FPN-backbone) for 24k iterations, while we fine-tune a FCN-16s for 90k iterations on both \texttt{train} sets and evaluate on the corresponding \texttt{val} sets for Cityscapes and BDD100K semantic segmentation.

%%%%%%%%%%%%%%%%%%%%%%%%%%%%%%%%%%%%%%%%%%%%%%%

\paragraph{Discussion.}

As shown in Table~\ref{tab:imagenet}, \ref{tab:coco} and \ref{tab:coco_2}, although originally designed for multi-instance circumstances, our {\it MultiSiam} still achieve state-of-the-art performance on all downstream benchmarks under different settings after pre-trained on ImageNet, revealing the generalization ability of {\it MultiSiam}.
The improvement is more significant when fine-tuning for 12k iterations compared with 90k and 180k settings, indicating that {\it MultiSiam} can effectively fasten model convergence.
Also the robustness to hyper-parameters will decrease the deployment difficulty to other domain-specific circumstances like medical images.  

%------------------------------------------------------------

\section{More Ablation Studies}\label{sec:more_ablation}

\subsection{Ablations on Optimization}

\paragraph{Minimum crop scale.}
During data augmentation, we will first select the scale of the output crop by randomly choosing a percentage of the original image scale between a minimum value and 100\%, followed by random cropping and flipping.
The results show that it's more beneficial to use a smaller minimum value to capture scale-invariance for downstream visual tasks.

\paragraph{Base learning rate.}
We find the optimal base learning rate for downstream dense visual tasks is much larger than that for image classification in BYOL, which demonstrates the differences between image-level visual tasks and pixel-level visual perception problems.

\paragraph{Base momentum.}
As we can see in Table~\ref{tab:optim}(c), using a larger base momentum and a more stable target network will help increase the transfer performance of the learned visual representations.

%------------------------------------------------------------

\subsection{Ablations on Self-attention Module}

We verify whether to use the residual connections in the self-attention mechanism of {\it MultiSiam} as the original Non-local Network~\cite{wang2018non} in Table~\ref{tab:selfatt}.
{\it RoI alignment} works slightly better with residual connections, while {\it offset alignment} suffers from significant performance drop.
We argue that using residual connections with {\it offset alignment} might hurt the prediction effectiveness under long range offset circumstances.

%------------------------------------------------------------

%%%%%%%%%%%%%%%%%%%%%%%%%%%%%%%%%%%%%%%%%%%%%%%

\begin{table}[t]
\begin{center}
\begin{tabular}{c|c|c|c}
\hline
Minimum  & Base & Base & \multirow{2}*{mIOU} \\
Crop Scale & Learning Rate & Momentum \\
\hline
\multicolumn{4}{{l}}{(a) Minimum Crop Scale} \\
\hline
0.08 & 0.3 & 0.99 & {\bf 71.9} \\
0.2 & 0.3 & 0.99 & 71.7 \\
\hline
\multicolumn{4}{{l}}{(b) Base Learning Rate} \\
\hline
0.08 & 0.3 & 0.99 & 71.9 \\
0.08 & 1.0 & 0.99 & {\bf 72.9} \\
\hline
\multicolumn{4}{{l}}{(c) Base Momentum} \\
\hline
0.2 & 0.1 & 0.99 & 71.6 \\
0.2 & 0.1 & 0.996 & {\bf 72.3} \\
\hline
\end{tabular}
\end{center}
\vspace{-2mm}
\caption{
{\bf Ablations on optimization hyper-parameters.} 
(a) Minimum crop scale; (b) base learning rate; (c) base momentum.
All results are evaluated on Cityscapes {\it val} set over three independent trials, same as the main paper.}
\vspace{-2mm}
\label{tab:optim}
\end{table}

%%%%%%%%%%%%%%%%%%%%%%%%%%%%%%%%%%%%%%%%%%%%%%%

\begin{table}[t]
\begin{center}
\begin{tabular}{c|c|c}
\hline
Feature Alignment  & Residual Connection & mIOU \\
\hline
RoI &  & 71.2 \\
RoI & \checkmark & 71.4 \\
Offset &  & {\bf 71.9} \\
Offset & \checkmark & 70.0 \\
\hline
\end{tabular}
\end{center}
\vspace{-2mm}
\caption{{\bf Ablations on residual connections in self-attention.}
As we can see, residual connections work well with {\it RoI alignment} but bring performance drop for {\it offset alignment}.}
\vspace{-2mm}
\label{tab:selfatt}
\end{table}

%%%%%%%%%%%%%%%%%%%%%%%%%%%%%%%%%%%%%%%%%%%%%%%

%------------------------------------------------------------

{\small
\bibliographystyle{ieee_fullname}
\bibliography{egbib.bib}
}

%------------------------------------------------------------

\end{document}